\documentclass[runningheads]{llncs}

\usepackage[T1]{fontenc}
\usepackage[utf8]{inputenc}

\usepackage{graphicx}
\usepackage{booktabs}
\usepackage{multirow}
\usepackage{tabularx}
\usepackage{makecell}
\usepackage{multicol}
\usepackage{orcidlink}
\usepackage{colortbl}
\usepackage{amsmath,amssymb,amsfonts}
\usepackage{xcolor}
\usepackage{microtype}
\usepackage{hyperref}
\hypersetup{
    hidelinks
}

\newcommand{\best}[1]{\textbf{#1}}
\newcommand{\secondbest}[1]{\underline{#1}}
\definecolor{embfull}{RGB}{255,232,218}
\definecolor{embfast}{RGB}{255,242,230}
\AddToHook{env/table/begin}{%
    \setlength{\belowcaptionskip}{4pt}%
    \setlength{\tabcolsep}{3pt}%
    \renewcommand{\arraystretch}{1.12}%
}
\AddToHook{env/figure/begin}{\setlength{\abovecaptionskip}{4pt}}
\AddToHook{env/figure/end}{\vspace{-3pt}}
\AddToHook{env/table/end}{\vspace{-3pt}}
\AddToHook{cmd/section/before}{\vspace{-6pt}}
\AddToHook{cmd/subsection/before}{\vspace{-5pt}}

\title{RecallGS: Geometry-Grounded Recall for Online Semantic Gaussian Splatting}

\titlerunning{RecallGS}

\author{Thanh-Khoi Nguyen\inst{1,2}\orcidlink{0009-0003-5879-451X}
\and
Thien-Phuc Tran \inst{1, 2}\orcidlink{0009-0008-0800-6884}
\and
Minh-Triet Tran \thanks{Corresponding author} \inst{1,2}\orcidlink{0000-0003-3046-3041}}
\authorrunning{Thanh-Khoi Nguyen}

\institute{University of Science, Ho Chi Minh City, Vietnam \and
Viet Nam National University, Ho Chi Minh City, Vietnam \\
\email{23120009@student.hcmus.edu.vn, ttphuc@selab.hcmus.edu.vn, tmtriet@fit.hcmus.edu.vn}
}

\begin{document}
\maketitle

\begin{abstract}
Online semantic 3D Gaussian Splatting (3DGS) enables open-vocabulary scene understanding from streaming observations. However, existing methods recover geometric context missing from image-derived semantic features using an additionally learned 3D network, incurring extra training and runtime overhead. We observe that such geometric-semantic evidence may already exist in the agent's observation history and can be retrieved rather than learned. Based on this insight, we introduce RecallGS (Geometry-Grounded Recall for Online Semantic Gaussian Splatting), a lightweight, training-free mechanism that treats past observations as a non-parametric geometric-semantic prior. RecallGS follows a recall-ground-fuse pipeline: CLIP retrieves relevant observations, alpha-compositing weights identify geometrically supported views, and aggregate their semantic features over each Gaussian's rendered footprint. On ScanNet and ScanNet++, RecallGS achieves strong semantic segmentation performance at 2.48 FPS, demonstrating that geometric-semantic context can be effectively recovered from observation history without an additional learned 3D branch.
\keywords{3D Gaussian Splatting, 3D Segmentation, Scene Understanding}
\end{abstract}

\section{Introduction}
3D Gaussian Splatting (3DGS) provides an efficient explicit representation for real-time novel-view synthesis\cite{kerbl20233d}. Recent works extend 3DGS with open-vocabulary semantics by distilling vision-language features into Gaussians, either through rendering-based supervision\cite{qin2024langsplat,shi2024legaussians,zhou2024feature3dgs} or direct 3D feature lifting and referring\cite{wu2024opengaussian,kim2025drsplat}. This paradigm has further expanded to online embodied settings, using either language-aware 3DGS-SLAM\cite{katragadda2025onlinelangsplat,zhou2025ea3d} or generalizable feed-forward reconstruction\cite{lee2026embodiedsplat}, alongside online 3D instance segmentation methods that fuse 2D foundation-model predictions into persistent 3D representations\cite{xu2024embodiedsam,tang2025onlineanyseg}.

Despite this progress, existing online semantic 3DGS pipelines recover the geometric context missing from image-derived language features through an additional learned 3D branch, rather than directly reusing geometric-semantic evidence from past observations. EmbodiedSplat\cite{lee2026embodiedsplat}, for instance, complements its 2D CLIP features \cite{radford2021learning} with geometry-aware 3D features produced by passing the local Gaussian point cloud through a sparse 3D U-Net\cite{choy20194d} equipped with a temporal memory adapter\cite{xu2024memoryadapters}. This 3D branch, however, must be trained on large-scale posed 3D indoor scene data, which is costly to collect and introduces additional GPU-memory consumption and latency that directly reduce the online reconstruction speed. EmbodiedSplat-fast removes the 3D U-Net and memory adapter, increasing throughput from 0.75 FPS to 5.18 FPS at the cost of geometric-semantic context. This highlights the training and inference overhead of recovering explicit 3D context through an additional learned branch. We ask whether such context must be learned at all, or can instead be recovered from the agent's observation history.

We observe that this trade-off is a consequence of treating each observation independently. An embodied agent operates in a persistent environment: a Gaussian that is geometrically or semantically ambiguous from the current viewpoint may correspond to a physical surface already observed earlier under a more favorable viewpoint, illumination, or occlusion pattern. This suggests that part of the geometric prior EmbodiedSplat's 3D branch must learn from parameters could instead be recovered from the agent's own trajectory, treating its accumulated observation history as a non-parametric geometric-semantic prior queried on demand rather than compressed once and discarded. However, effectively reusing past observations introduces several challenges. First, semantic similarity does not imply physical correspondence: a CLIP-based retrieval system\cite{radford2021learning} may return frames containing a visually similar object-another chair, table, or door-at an entirely different location. Second, a Gaussian may project inside a historical image but still be occluded by closer geometry. Sampling the projected pixel would then transfer features from the occluder rather than from the queried surface. 
Finally, exhaustively verifying every historical camera against every incoming frame would itself undermine the efficiency this history is meant to provide, as a long-running agent may accumulate hundreds to thousands of observations. The problem is therefore not retrieval of similar-looking memories, but memory routing: determining, without prior semantic knowledge of the current Gaussians, which past observations should be allowed to influence each one.

To this end, we propose \textbf{RecallGS} (Geometry-Grounded Recall), a lightweight mechanism requiring no additional training that treats an agent's observation history as a queryable source of geometric-semantic evidence, routed to the current Gaussians through a recall–ground–replay pipeline. Our contributions are threefold: (1) semantic recall, a cheap CLIP-based retrieval stage that narrows an unbounded observation history to a small candidate pool of appearance-plausible frames; (2) geometric grounding, which reranks these candidates using alpha-compositing weights to identify views that geometrically support the current Gaussians; and (3) compositing fusion, which reuses the same alpha-compositing weights to aggregate historical features over each Gaussian's full rendered footprint. Our experiments show that each component contributes independently to semantic accuracy, and that their combination enables effective geometric-semantic reasoning from observation history while maintaining efficient online inference

Our contributions are as follows:
\begin{itemize}
\item We show that an agent's observation history can serve as a non-parametric geometric-semantic prior, providing complementary semantic evidence without additional 3D training or per-scene optimization.
\item We propose \textbf{RecallGS}, a training-free recall-ground-fuse framework that efficiently retrieves relevant past observations, grounds them against the current geometry, and fuses their semantic evidence into the current Gaussians.
\item We validate the contribution of each component and show that exploiting observation history consistently improves online semantic 3DGS while maintaining efficient inference.
\end{itemize}

\section{Related Work}
\subsection{Language-Embedded Gaussian Splatting}
Building on 3D Gaussian Splatting\cite{kerbl20233d}, one line of work embeds
open-vocabulary semantics onto Gaussians via per-scene optimization, either by
distilling 2D vision-language features through differentiable rendering
supervision\cite{qin2024langsplat,shi2024legaussians,zhou2024feature3dgs} or
by directly associating semantic features with 3D structure through clustering,
feature lifting, or compact Gaussian-level representations~
\cite{wu2024opengaussian,kim2025drsplat,li2025instancegaussian,cheng2025occamlgs,cen2025saga,qiu2024gls,shorinwa2024fastsplat}. These methods establish 3DGS as an effective carrier of language features and improve their spatial consistency, instance awareness, or representation efficiency. However, most construct semantic fields specifically for each scene, which limits their applicability when observations arrive continuously. 

To remove scene-specific optimization, a class of feed-forward methods predicts 3D Gaussian representations directly from input images without per-scene training. This paradigm was established by geometric backbones such as pixelSplat\cite{charatan2024pixelsplat} and MVSplat\cite{chen2024mvsplat}, which regress RGB Gaussians from sparse image pairs. Building on this feed-forward formulation, subsequent work adds an open-vocabulary semantic branch: LSM \cite{fan2024lsm} jointly regresses geometry, appearance, and semantics from unposed image pairs by lifting a pre-trained 2D language-based segmentation model into a 3D-consistent feature field decoded into semantic Gaussians, SLGaussian \cite{chen2025slgaussian} targets the sparse-view regime with a multi-view language memory bank that maps high-dimensional CLIP features to compact per-object IDs for fast open-vocabulary querying, and SemanticSplat \cite{li2025semanticsplat} further
improves language-aware feed-forward reconstruction by fusing SAM~\cite{kirillov2023segment} and CLIP-LSeg~\cite{li2022language} features into a cost-volume representation for cross-view semantic aggregation. These approaches avoid per-scene semantic optimization and generalize across scenes, but still operate on a fixed, small set of input views rather than maintaining a persistent semantic representation under a streaming observation sequence.

\subsection{Online Gaussian Splatting Reconstruction and Segmentation}
A separate line of work extends 3DGS to the online setting required by embodied perception. Within visual SLAM, SplaTAM~\cite{keetha2024splatam} and Gaussian Splatting SLAM~\cite{matsuki2024gsslam} use explicit Gaussians for incremental tracking, mapping, and rendering, demonstrating that 3DGS can serve as a persistent scene representation under streaming observations. Building on this formulation, Online-LangSplat~\cite{katragadda2025onlinelangsplat} and EA3D~\cite{zhou2025ea3d} inject language or foundation-model features into the online reconstruction loop. Online-LangSplat's language module itself runs in milliseconds, but the overall SLAM-GS pipeline remains bottlenecked by per-frame camera-pose optimization, yielding well under 2 FPS end-to-end; EA3D similarly
relies on a recurrent joint optimization module to align geometric and semantic features, at a comparable throughput cost. In parallel, online 3D instance segmentation methods incrementally fuse 2D foundation-model predictions into temporally consistent 3D instances~\cite{xu2024embodiedsam,tang2025onlineanyseg}, building on adapter-based caching mechanisms that aggregate RGB-D features frame-by-frame to endow otherwise-offline backbones with temporal learning ability~\cite{xu2024memoryadapters}; these methods, however, operate on
point-based representations rather than a differentiable and renderable Gaussian semantic field.

Closest to our setting are online, generalizable feed-forward pipelines that
avoid per-scene optimization entirely. FreeSplat++~\cite{wang2025freesplat++} targets efficient whole-scene indoor reconstruction from long image sequences, using a low-cost cross-view aggregation module and a pixel-wise triplet fusion strategy to incrementally merge overlapping Gaussian primitives across views, together with a weighted floater-removal step that exploits the accumulated blending weights from this fusion. EmbodiedSplat~\cite{lee2026embodiedsplat} extends this backbone with open-vocabulary semantics through a sparse CLIP codebook and a learned 3D U-Net branch that injects geometric context into image-derived semantic features. This improves geometric-semantic awareness but requires additional 3D training and incurs non-negligible inference overhead. In contrast, we recover complementary geometric-semantic evidence directly from the agent's own observation history and route it to the current Gaussians through a lightweight recall-ground-fuse mechanism without an additional learned 3D semantic branch.
\section{Method}
\subsection{Preliminaries}
\label{sec:preliminaries}

\noindent\textbf{3D Gaussian Splatting}
3DGS \cite{kerbl20233d} represents a scene as Gaussian primitives
$\Theta=\{\mu_i,\Sigma_i,\alpha_i,\mathbf{c}_i\}_{i=1}^{M}$, rendered along each ray by
\begin{equation}
\hat{\mathbf{c}} = \sum_{i=1}^{N_r} T_i \tilde\alpha_i \mathbf{c}_i, \qquad
\tilde\alpha_i = \alpha_i \exp\!\left(-\tfrac{1}{2}\mathbf{d}^\top\Sigma_{2D}^{-1}\mathbf{d}\right),
\end{equation}
where $N_r$ is the number of contributing Gaussians, $T_i$ the transmittance, $\Sigma_{2D}$ the projected covariance, and $\mathbf{d}$ the displacement from the projected mean. For pixel $p$ in camera $j$, we use the transmittance $T_{ip}^{(j)}$ and effective opacity $\tilde\alpha_{ip}^{(j)}$. Their product,
$w_{ip}^{(j)}=T_{ip}^{(j)}\tilde\alpha_{ip}^{(j)}$, defines the rendering responsibility used for visibility reasoning in Sec.~\ref{sec:method-filter}.

\textbf{FreeSplat++}
Following EmbodiedSplat~\cite{lee2026embodiedsplat}, we adopt
FreeSplat++~\cite{wang2025freesplat++} as the feed-forward reconstruction backbone. At timestep $t$, FreeSplat++ extracts multi-view image features, aggregates them through its cross-view reconstruction pipeline, and predicts pixel-aligned local Gaussian triplets
$\Theta_l^t=\{\mu_l^t,\gamma_l^t,\mathbf{f}_l^t\}$, where $\mu_l^t$, $\gamma_l^t$, and $\mathbf{f}_l^t$ denote the Gaussian centers, fusion confidences, and latent features, respectively. These triplets form a latent Gaussian representation rather than the full rendering parameters $\{\mu,\Sigma,\alpha,\mathbf{c}\}$. Overlapping local and global
triplets are incrementally fused through confidence-weighted center updates and GRU-based latent fusion. FreeSplat++ decodes the fused latent representation into rendering parameters after aggregation; for geometric grounding, we apply the same decoder to the current local triplets at every timestep to obtain covariance, opacity, and color. We denote the resulting renderable local Gaussians by $\hat{\Theta}_l^t$, which preserve the centers and indices of $\Theta_l^t$.

\textbf{EmbodiedSplat}
EmbodiedSplat \cite{lee2026embodiedsplat} provides open-vocabulary semantics for this online backbone. Since storing a dense CLIP feature for every Gaussian is prohibitive at scene scale ($M$ often exceeds $10^6$), it maintains a \emph{CLIP Global Codebook} $\mathbf{C}^t$ of instance-level features,
$\mathbf{C}^t=\mathrm{concat}(\mathbf{C}^{t-1},\hat{\mathbf{s}}^t)$, together with a fixed-length per-Gaussian \emph{sparse coefficient field}
$\{\mathrm{Idx}_g^t,\Omega_g^t\}$ that stores $L$ codebook indices and coefficients instead of a dense $D$-dimensional feature. The semantic feature of Gaussian $i$ is reconstructed as
\begin{equation}
\mathbf{s}_g^{T_{\mathrm{end}}}(i)
=
\sum_{k=1}^{L}
\Omega_g^{T_{\mathrm{end}}}(i,k)\,
\mathbf{C}^{T_{\mathrm{end}}}\!\big(\mathrm{Idx}_g^{T_{\mathrm{end}}}(i,k)\big),
\qquad
\sum_{k=1}^{L}\Omega_g^{T_{\mathrm{end}}}(i,k)=1.
\end{equation}
Because the codebook stores unmodified CLIP vectors rather than a learned bottleneck, it requires neither per-scene optimization nor pretraining, while retaining CLIP's open-vocabulary generalization and supporting online updates through fixed-length caches. For semantic replay, we additionally cache a dense pixel-aligned CLIP feature map $F_j(p)\in\mathbb{R}^{D}$ for each historical frame $I^j$, distinct from the instance-pooled features $\hat{\mathbf{s}}^j$.
\subsection{Problem Formulation}
\label{sec:problem_formulation}

At timestep $t$, the system receives a posed RGB observation $\mathcal{O}^t$ and maintains a scene state $\mathcal{S}^t$:
\begin{equation}
    \mathcal{O}^t = \left(I^t, \mathbf{K}^t, \mathbf{E}^t\right), \qquad
    \mathcal{S}^t = \left(\hat{\Theta}_g^t, \mathbf{C}^t, \mathrm{Idx}_g^t, \Omega_g^t\right).
\end{equation}
Here, $I^t$ is the current RGB frame, $\mathbf{K}^t$ the camera intrinsics, and $\mathbf{E}^t$ the camera-to-world pose. Following EmbodiedSplat, $\hat{\Theta}_g^t$ is the global Gaussian map, incrementally updated from the current local prediction $\hat{\Theta}_l^t$, while the semantic state comprises the CLIP Global Codebook $\mathbf{C}^t$ and the per-Gaussian sparse coefficient field $\{\mathrm{Idx}_g^t, \Omega_g^t\}$. We denote by $g_i^t\in\hat\Theta_l^t$ the $i$-th local Gaussian at timestep $t$, represented by its decoded parameters when rasterized. Given the streaming sequence $\mathcal{O}^{1:t}$, our goal is to construct this persistent geometric and open-vocabulary semantic representation online. RecallGS additionally maintains a fixed-size episodic memory
\begin{equation}
    \mathcal{H}^{t-1}_W =
    \left\{
    (I^j,\mathbf{K}^j,\mathbf{E}^j,q^j,F_j)
    \right\}_{j=\max(1,t-W)}^{t-1},
\end{equation}
containing at most $W$ recent observations. Each entry stores the RGB frame, camera calibration and pose, image-level retrieval descriptor $q^j$, and pixel-aligned feature map $F_j$. Once the memory reaches capacity, the oldest observation is discarded. The online state update is therefore
\begin{equation}
    \mathcal{S}^t =
    \mathcal{U}\left(
    \mathcal{O}^t,
    \mathcal{S}^{t-1},
    \mathcal{H}^{t-1}_W
    \right).
    \label{eq:online_problem}
\end{equation}
The model must produce $\mathcal{S}^t$ using only $\mathcal{O}^{1:t}$, without access to future frames or offline per-scene optimization.

\subsection{Overview}
\begin{figure}[t]
    \centering
    \includegraphics[width=1\linewidth]{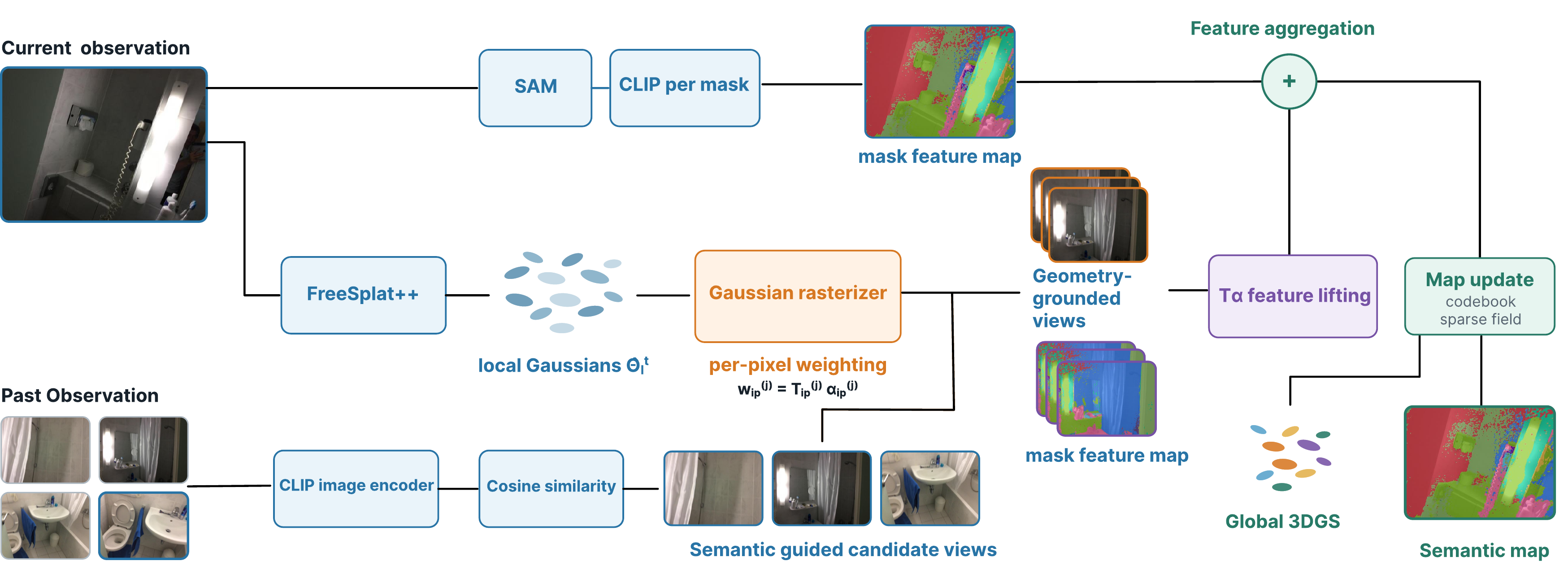}
    \caption{Overview of RecallGS. Given a posed observation and the observation history, RecallGS recalls candidate views via CLIP, verifies their geometric support with rendering responsibility, and replays the verified features to build an online 3DGS reconstruction with a semantic field}
    \label{fig:pipeline_embodied_splat}
\end{figure}

As shown in Fig~\ref{fig:pipeline_embodied_splat}, RecallGS equips a lightweight online semantic mapping pipeline with episodic memory. For each incoming observation, we retrieve relevant past views, verify their geometric compatibility with the current scene, and reuse their semantic evidence to refine the current representation. This enables the model to exploit previously observed information while preserving an efficient online pipeline.

\subsection{Semantic Recall over Episodic Memory}
\label{sec:method-retrieval}

EmbodiedSplat grounds created Gaussians $g_i^t \in \hat{\Theta}_l^t$ using semantic features from the current observation, while its global map and sparse coefficient field propagate fused semantics across timesteps. However, it does not explicitly reuse per-frame semantic features from past observations for newly introduced Gaussians. RecallGS addresses this limitation by querying its episodic memory $\mathcal{H}^{t-1}_W$. A newly introduced Gaussian may correspond to a surface observed earlier under a more favorable viewpoint, illumination, or occlusion pattern; retrieving such observations provides complementary semantic evidence for the current local Gaussians.

Although the episodic memory is bounded by $W$, only a small subset of its observations is typically relevant to the current local Gaussians. Geometrically evaluating all cached observations would therefore require unnecessary rasterization from camera poses unlikely to provide useful evidence. We adopt a coarse-to-fine strategy that first uses inexpensive appearance-based retrieval to select a small set of plausible observations from $\mathcal{H}^{t-1}_W$, and applies geometry-dependent verification only to the recalled candidates (Sec.~\ref{sec:method-filter}).

Concretely, in addition to the per-instance CLIP feature $\hat{\mathbf{s}}^t$ that EmbodiedSplat extracts from $I^t$ to populate the codebook $\mathbf{C}^t$ (Sec.~\ref{sec:preliminaries}), we compute a global image-level CLIP embedding
\begin{equation}
    q^t = \mathrm{CLIP}(I^t) \in \mathbb{R}^{D}.
\end{equation}
The two features serve different purposes: $\hat{\mathbf{s}}^t$ is an instance-level feature associated with a segmentation mask and used to construct per-Gaussian semantics, whereas $q^t$ is a single image-level descriptor used only for retrieving observations from the episodic memory. Each $q^j$ is computed once when its observation enters the memory and remains fixed while that observation is retained.

Given the current embedding $q^t$, we compute its cosine similarity with each observation retained in $\mathcal{H}^{t-1}_W$ and retrieve the $B_{\mathrm{rec}}$ highest-scoring observations:
\begin{equation}
    \rho_j^{\mathrm{clip}}
    =
    \frac{(q^t)^\top q^j}
    {\lVert q^t\rVert_2\,\lVert q^j\rVert_2},
    \qquad
    \mathcal{R}^t
    =
    \operatorname*{Top\text{-}B_{\mathrm{rec}}}_{
        \max(1,t-W)\leq j<t
    }
    \rho_j^{\mathrm{clip}}.
    \label{eq:candidate-retrieval}
\end{equation}
Here, $B_{\mathrm{rec}}\ll W$ is the recall budget, and $\operatorname{Top\text{-}B}$ returns all available observations when fewer than $B$ are stored. The resulting candidate set $\mathcal{R}^t$ restricts the more expensive geometric verification to at most $B_{\mathrm{rec}}$ observations, as described next.

\subsection{Geometric Grounding of Recalled Memories}
\label{sec:method-filter}

The candidate pool $\mathcal{R}^t$ obtained in Sec.~\ref{sec:method-retrieval} is constructed entirely from image-level appearance and carries no guarantee about the underlying 3D scene. Two frames may be assigned a high similarity score $\rho_j^{\mathrm{clip}}$ simply because they share dominant visual attributes, such as similar wall color or furniture style, without observing any common physical surface; conversely, a historical camera $\mathbf{E}^j$ that does observe the region corresponding to $\hat\Theta_l^t$ may receive only a modest score if the remainder of $I^j$ is dominated by unrelated content. Using $\mathcal{R}^t$ directly as historical evidence would therefore risk contaminating the current Gaussians with semantics harvested from frames that never actually depicted the queried geometry. We resolve this ambiguity by explicitly verifying, for every candidate $j \in \mathcal{R}^t$, how much of $\hat\Theta_l^t$ it renders once projected through the true camera geometry, rather than relying on appearance alone.

A naive verification would rasterize $\hat\Theta_l^t$ from camera $j$ and treat a Gaussian as visible whenever its projected center falls within the image bounds, or equivalently whenever its screen-space radius is non-zero. Such a criterion, however, only certifies that Gaussian $i$ possesses a footprint in the image plane; it says nothing about whether that footprint survives occlusion by other Gaussians nearer to camera $j$ along the same ray. Under the alpha-compositing model of Sec.~\ref{sec:preliminaries}, a Gaussian may lie well within the frustum of camera $j$ and still contribute negligibly to the rendered image if its transmittance $T_{ip}^{(j)}$ has already been suppressed by closer geometry, in which case treating it as "visible" would inject semantic evidence from a view that, for that Gaussian, is effectively occluded. We therefore reject binary, projected-center visibility in favor of a criterion grounded directly in the rendering weight $w_{ip}^{(j)} = T_{ip}^{(j)}\tilde\alpha_{ip}^{(j)}$, which already accounts for occlusion among the rasterized Gaussians, partial coverage, and the soft falloff of the Gaussian kernel.

Concretely, for each candidate camera $j \in \mathcal{R}^t$, we rasterize the current local Gaussian set $\hat\Theta_l^t$ through $(\mathbf{K}^j, \mathbf{E}^j)$ and accumulate, for every Gaussian $g_i^t \in \hat\Theta_l^t$, its total rendering contribution across the image plane,
\begin{equation}
    A_i^{(j)} = \sum_{p} w_{ip}^{(j)}.
\end{equation}
A Gaussian is considered geometrically supported by camera $j$ if its accumulated contribution exceeds a fixed threshold $\tau$, and we define the coverage of camera $j$ with respect to the current local Gaussian set as the fraction of Gaussians satisfying this criterion,
\begin{equation}
    \kappa_j = \frac{1}{N_t}\sum_{i=1}^{N_t} \mathbf{1}\!\left[A_i^{(j)} > \tau\right], \qquad N_t = \lvert\hat\Theta_l^t\rvert.
    \label{eq:coverage}
\end{equation}
Intuitively, $\kappa_j$ answers a question that $\rho_j^{\mathrm{clip}}$ cannot: not merely whether camera $j$ depicts similar content to $I^t$, but what fraction of the geometry newly introduced at timestep $t$ was actually, and non-negligibly, observed by camera $j$.
Because $A_i^{(j)}$ is an unnormalized sum over pixels, the fixed threshold $\tau$ depends on image resolution and projected footprint size. It therefore measures visible rendering mass jointly with footprint size, rather than visibility alone.

With a fixed grounding budget $B_{\mathrm{geo}}$, we select the candidates with the highest coverage as the final historical evidence set,
\begin{equation}
    \mathcal{V}^t = \operatorname*{Top\text{-}B_{\mathrm{geo}}}_{j \in \mathcal{R}^t} \; \kappa_j, \qquad B_{\mathrm{geo}} \ll B_{\mathrm{rec}}.
    \label{eq:coverage-selection}
\end{equation}
Combining Eqs.~\eqref{eq:candidate-retrieval} and~\eqref{eq:coverage-selection}, the selection process is
\begin{equation}
    \mathcal{H}^{t-1}_W
    \xrightarrow{\ \text{Semantic Recall}\ }
    \mathcal{R}^t
    \xrightarrow{\ \text{Geometric Grounding}\ }
    \mathcal{V}^t.
\end{equation}
Semantic recall first reduces the full history to at most $B_{\mathrm{rec}}$ plausible candidates, after which geometric grounding retains only the views that provide verified support for the current Gaussians. This avoids rasterizing the entire history while ensuring that the selected views are geometrically relevant to $\hat{\Theta}_l^t$.

\subsection{Compositing Fusion into the Global Codebook}
\label{sec:method-integration}

We realize semantic replay via responsibility-weighted feature aggregation over the per-Gaussian historical view set $\mathcal{V}_i^t$ defined below. Rendering-weighted multi-view feature aggregation has also been explored in language Gaussian splatting\cite{cheng2025occamlgs}; our use of the same rendering responsibility is instead applied to causal historical replay in an online feed-forward system. The output of the two modules above, $\mathcal{V}^t$, identifies historical cameras verified to observe $\hat\Theta_l^t$, but must still be reintroduced at the point where EmbodiedSplat grounds each Gaussian from $I^t$ alone. For each Gaussian $g_i^t$, we restrict evidence to views satisfying the same coverage criterion, $\mathcal{V}_i^t = \{j \in \mathcal{V}^t : A_i^{(j)} > \tau\}$, and sample the cached dense feature map $F_j(\cdot)$ over its rendered footprint, weighted by $w_{ip}^{(j)}$:
\begin{equation}
    \mathbf{s}_{i,j}^{\mathrm{hist}} = \frac{\sum_p w_{ip}^{(j)} F_j(p)}{A_i^{(j)} + \epsilon}, \qquad
    \mathbf{s}_i^{\mathrm{hist}} = \frac{\sum_{j \in \mathcal{V}_i^t} A_i^{(j)}\, \mathbf{s}_{i,j}^{\mathrm{hist}}}{\sum_{j \in \mathcal{V}_i^t} W_i^{(j)} + \epsilon}.
\end{equation}
Here $\epsilon>0$ stabilizes the denominators. The current-frame feature $\mathbf{s}_i^{\mathrm{cur}}\in\mathbb{R}^{D}$ is obtained by the baseline current-frame grounding: the per-Gaussian sparse coefficients weight the current instance-level CLIP entries $\hat{\mathbf{s}}^t$ assigned to Gaussian $i$, and the resulting weighted sum is $\ell_2$-normalized. It lies in the same CLIP feature space as the codebook entries and $\mathbf{s}_g^{T_{\mathrm{end}}}(i)$, but precedes global fusion. We fuse the historical and current-frame features as
\begin{equation}
    \mathbf{s}_i^{\mathrm{final}} = \operatorname{Norm}\!\left(\mathbf{s}_i^{\mathrm{cur}} + \lambda\, \mathbf{s}_i^{\mathrm{hist}}\right),
\end{equation}
where $\lambda$ controls the historical contribution and $\operatorname{Norm}$ denotes $\ell_2$ normalization. Since $\mathbf{s}_i^{\mathrm{cur}}$ is unit-normalized, this reduces to $\mathbf{s}_i^{\mathrm{final}} = \mathbf{s}_i^{\mathrm{cur}}$ when $\mathcal{V}_i^t = \varnothing$. $\mathbf{s}_i^{\mathrm{final}}$ is then passed to the sparse coefficient field and global codebook unchanged, so our two modules act purely as an upstream semantic evidence source, requiring no additional training.

\section{Experiment}
\textbf{Dataset:}
We follow the same dataset and evaluation setup as EmbodiedSplat\cite{lee2026embodiedsplat}. Specifically, we evaluate on the indoor ScanNet\cite{dai2017scannet} and ScanNet++\cite{yeshwanth2023scannetpp} benchmarks using the same preprocessed test scenes and ground-truth semantic annotations provided by the official EmbodiedSplat implementation. For ScanNet, we report 3D semantic segmentation under the 10, 15, and 19 class evaluation protocols, while ScanNet++ is evaluated under the 20-class setting. Unless otherwise stated, all dataset preprocessing, scene splits, and evaluation protocols are kept identical to EmbodiedSplat to ensure direct comparability.

\textbf{Baselines.}
We compare RecallGS against representative open-vocabulary semantic 3DGS methods, including OpenGaussian\cite{wu2024opengaussian}, Occam's LGS\cite{cheng2025occamlgs}, Dr.Splat\cite{kim2025drsplat}, and InstanceGaussian\cite{li2025instancegaussian}, as well as both \emph{EmbodiedSplat} and \emph{EmbodiedSplat-fast}\cite{lee2026embodiedsplat}. Among them, EmbodiedSplat is the most relevant reference, as it augments the feed-forward reconstruction backbone with an additional geometry-aware 3D semantic branch, while EmbodiedSplat-fast removes this branch and relies only on the lightweight 2D semantic pathway. Since RecallGS is built upon EmbodiedSplat-fast, we use it as the primary baseline to isolate the benefit of historical semantic replay, while comparison with the full EmbodiedSplat evaluates whether historical evidence can recover geometry-aware semantic context without the additional 3D U-Net.

\textbf{Implementation Details.}
We implement RecallGS on top of the official EmbodiedSplat-fast, keeping the FreeSplat++ backbone and semantic codebook unchanged. Unless otherwise specified, we use $B_{\mathrm{rec}}=16$, $B_{\mathrm{geo}}=3$, $\tau=0.9$, and $\lambda=0.75$, following the replay procedure described in Sec.~3. We evaluate semantic segmentation using mIoU and mAcc on ScanNet-10/15/19 and ScanNet++-20, and report FPS under the same hardware and evaluation setting. All experiments are conducted in a single NVIDIA RTX5090 GPU.

\textbf{Quantitative result}
Table~\ref{tab:main-comparison} shows that RecallGS improves EmbodiedSplat-fast by $3.10$, $3.19$, $3.38$, and $3.22$ mIoU on ScanNet-10/15/19 and ScanNet++-20, respectively, with corresponding mAcc gains. It remains competitive with the full model while running at $2.48$ FPS versus $0.75$ FPS ($\approx 3\times$ faster). Although slower than EmbodiedSplat-fast ($5.18$ FPS), RecallGS provides a favorable accuracy--efficiency trade-off and compares favorably with per-scene methods while retaining online inference.

\begin{table*}[!t]
\centering
\scriptsize
\caption{Main comparison on ScanNet and ScanNet++. We report mIoU, mAcc, and online FPS; a dash denotes offline methods. Best and second-best results are \textbf{bold} and \underline{underlined}, respectively.}
\label{tab:main-comparison}
\setlength{\tabcolsep}{2.4pt}
\renewcommand{\arraystretch}{1.05}
\resizebox{\textwidth}{!}{%
\begin{tabular}{l|rr|rr|rr||rr|r|l|l}
\toprule
Method & \multicolumn{2}{c|}{ScanNet-10} & \multicolumn{2}{c|}{ScanNet-15} & \multicolumn{2}{c||}{ScanNet-19} & \multicolumn{2}{c|}{ScanNet++-20} & Avg FPS & Scope & Mode \\
& mIoU & mAcc & mIoU & mAcc & mIoU & mAcc & mIoU & mAcc & & & \\
\midrule
LangSplat & 6.52 & 20.11 & 3.25 & 13.16 & 1.34 & 7.44 & 2.21 & 10.18 & - & \multirow{3}{*}{Per-Scene} & \multirow{2}{*}{Offline} \\
LEGaussians & 6.79 & 18.13 & 4.13 & 15.49 & 2.53 & 5.67 & 2.93 & 9.34 & - & & \\
Online-LangSplat & 7.13 & 21.56 & 3.89 & 14.52 & 3.45 & 8.97 & 4.51 & 11.34 & 1.12 & & Online \\
\midrule
OpenGaussian & 29.50 & 44.61 & 23.74 & 39.14 & 22.52 & 35.02 & 25.65 & 37.03 & - & \multirow{4}{*}{Per-Scene} & \multirow{4}{*}{Offline} \\
Occam's LGS & 42.14 & 70.28 & 35.04 & 63.71 & 30.49 & 57.91 & 34.08 & 61.19 & - & & \\
Dr. Splat & 39.21 & 66.66 & 31.84 & 60.58 & 28.38 & 55.85 & 39.85 & 58.34 & - & & \\
InstanceGaussian & 29.77 & 52.32 & 28.79 & 50.07 & 26.57 & 48.63 & 29.98 & 47.47 & - & & \\
\midrule
EmbodiedSplat (RGB) & \secondbest{49.81} & 76.13 & \best{49.23} & 75.47 & \best{46.22} & \secondbest{70.37} & 41.93 & 61.50 & 0.75 & \multirow{3}{*}{Generalizable} & \multirow{3}{*}{Online} \\
EmbodiedSplat-fast (RGB) & 47.86 & \secondbest{77.62} & 43.21 & 73.85 & 41.03 & 70.12 & \secondbest{45.53} & \secondbest{71.42} & \best{5.18} & & \\
\textbf{RecallGS} & \best{50.96} & \best{81.55} & \secondbest{46.40} & \best{76.97} & \secondbest{44.41} & \best{73.26} & \best{48.75} & \best{75.12} & \secondbest{2.48} & & \\
\bottomrule
\end{tabular}%
}
\end{table*}

\begin{figure}[!t]
\centering
\includegraphics[width=0.95\linewidth]{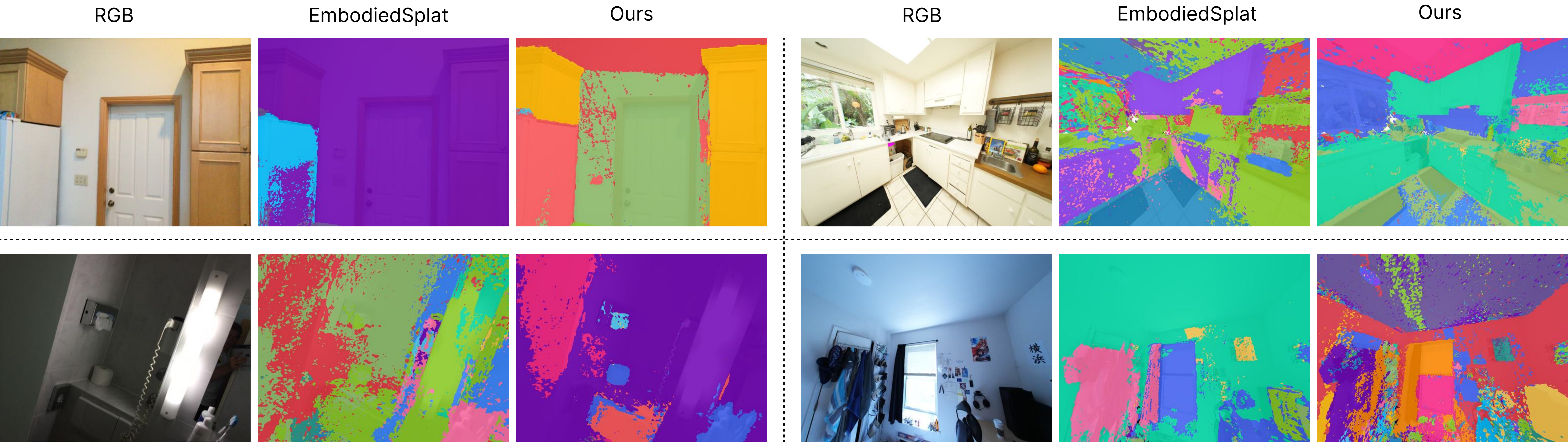}
\caption{Qualitative comparison on ScanNet and ScanNet++. RecallGS produces more coherent semantic regions and fewer artifacts than EmbodiedSplat-fast, particularly on large surfaces and partially observed boundaries.}\label{fig:qualitative_embodied_splat}
\end{figure}

\textbf{Qualitative result:}
Fig.~\ref{fig:qualitative_embodied_splat} demonstrates that RecallGS produces substantially more coherent semantic reconstructions, with cleaner region continuity and fewer fragmented or streak-like artifacts across large surfaces and object boundaries. In contrast, EmbodiedSplat exhibits pronounced local inconsistency, particularly in cluttered regions and partially observed structures. These results indicate that geometrically grounded historical replay provides complementary evidence that stabilizes semantic predictions across viewpoints and improves the consistency of the reconstructed semantic field.

\section{Ablation study}
\subsection{Historical View Selection}

Replacing temporally recent observations with CLIP-based retrieval improves performance across the four evaluation protocols in Table~\ref{tab:semantic-recall-ablation}, yielding gains of approximately $0.44$-$0.63$ mIoU. This indicates that temporal proximity alone is insufficient for identifying informative historical evidence, whereas semantic similarity provides a stronger prior for memory selection. Nevertheless, directly replaying the top-$B_{\mathrm{geo}}$ CLIP-retrieved views remains suboptimal. Incorporating Gaussian-based geometric reranking further improves performance by $0.54$-$0.88$ mIoU, demonstrating that semantic relevance does not necessarily imply geometric compatibility with the current reconstruction. Overall, the full recall-and-grounding strategy outperforms recent-history selection by approximately $1.14$-$1.51$ mIoU, validating the complementary roles of semantic retrieval and geometric verification in this study.

\begin{table}[!t]
\centering
\small
\caption{Ablation of historical-view retrieval}
\label{tab:semantic-recall-ablation}
\renewcommand{\arraystretch}{1.0}
\begin{tabular}{@{}lrrrr@{}}
\toprule
Variant & ScanNet-10 & ScanNet-15 & ScanNet-19 & ScanNet++-20 \\
\midrule
Recent History & 49.74 & 44.89 & 43.07 & 47.61 \\
CLIP Only & 50.18 & 45.52 & 43.57 & 48.21 \\
\best{Full (RecallGS)} & \best{50.96} & \best{46.40} & \best{44.41} & \best{48.75} \\
\bottomrule
\end{tabular}
\end{table}

\subsection{Compositing vs. Center-Based Fusion}
Table~\ref{tab:geometric-grounding-ablation} studies how Gaussian geometry is used during historical replay. \emph{Projected Center} uses only the projected Gaussian center for both historical-view ranking and semantic feature assignment. \emph{Center Lifting} retains our $T\alpha$-based coverage for selecting historical views, but assigns each Gaussian the semantic feature sampled only at its projected center. In contrast, the full model uses $T\alpha$ throughout: historical cameras are ranked using the accumulated Gaussian contribution, and semantic features are aggregated over the complete projected footprint using the same rendering weights. The full formulation improves over Projected Center by approximately $1.22$-$1.69$ mIoU and over Center Lifting by $0.69$-$0.91$ mIoU, indicating the expected benefit of exploiting the complete Gaussian footprint for both historical observability and semantic transfer. These variants do not isolate the transmittance factor $T$ from opacity $\alpha$; therefore, the occlusion-specific contribution of $T$ is not separately measured by this ablation.

\begin{table}[!t]
\centering
\small
\caption{Ablation on compositing vs. center-based fusion.}
\label{tab:geometric-grounding-ablation}
\renewcommand{\arraystretch}{1.0}
\begin{tabular}{@{}lrrrr@{}}
\toprule
Variant & ScanNet-10 & ScanNet-15 & ScanNet-19 & ScanNet++-20 \\
\midrule
Projected Center & 49.27 & 45.18 & 42.86 & 47.34 \\
Center Lifting & 50.05 & 45.62 & 43.58 & 48.06 \\
\best{Full $T\alpha$ (RecallGS)} & \best{50.96} & \best{46.40} & \best{44.41} & \best{48.75} \\
\bottomrule
\end{tabular}
\end{table}

\subsection{Recall and Grounding Budget Sensitivity}

\begin{table}[!t]
\centering
\scriptsize
\caption{Retrieval and replay budget sensitivity (default: $B_{\mathrm{rec}}=16$, $B_{\mathrm{geo}}=3$).}
\label{tab:mk_ablation}
\setlength{\tabcolsep}{2.5pt}
\renewcommand{\arraystretch}{1.0}
\begin{tabular*}{\linewidth}{@{\extracolsep{\fill}}crrr@{\hspace{8pt}}crrr@{}}
\toprule
\multicolumn{4}{c}{\textit{Varying $B_{\mathrm{rec}}$ ($B_{\mathrm{geo}}=3$)}} &
\multicolumn{4}{c}{\textit{Varying $B_{\mathrm{geo}}$ ($B_{\mathrm{rec}}=16$)}} \\
\cmidrule(r){1-4}\cmidrule(l){5-8}
$B_{\mathrm{rec}}$ & ScanNet-10 & ScanNet-15 & ScanNet-19 &
$B_{\mathrm{geo}}$ & ScanNet-10 & ScanNet-15 & ScanNet-19 \\
\midrule
8  & 50.82 & 46.12 & 44.02 &
2  & 50.71 & 46.08 & 43.91 \\
\best{16} & \best{50.96} & \best{46.40} & \best{44.41} &
\best{3} & \best{50.96} & \best{46.40} & \best{44.41} \\
32 & 50.77 & 46.05 & 43.96 &
5  & 50.84 & 46.17 & 44.05 \\
\bottomrule
\end{tabular*}
\end{table}

Table~\ref{tab:mk_ablation} varies the recall budget $B_{\mathrm{rec}}$ and the
grounding budget $B_{\mathrm{geo}}$. Performance is moderately sensitive to both:
moving away from the default ($B_{\mathrm{rec}}=16$, $B_{\mathrm{geo}}=3$) lowers
mIoU by 0.12-0.50 across the three ScanNet protocols, and every setting remains
above the CLIP-only and Center Lifting variants in
Tables~\ref{tab:geometric-grounding-ablation} and~2. A smaller $B_{\mathrm{rec}}$
may exclude informative views from the candidate pool, whereas a larger one
admits more appearance-similar but geometrically unsupported candidates.
Likewise, a smaller $B_{\mathrm{geo}}$ provides less complementary evidence, while
a larger one adds redundant or weakly supported views. We therefore use
$B_{\mathrm{rec}}=16$ and $B_{\mathrm{geo}}=3$ as the default; note that this
setting was chosen on the same evaluation scenes

\section{Conclusion}
We propose RecallGS, a lightweight mechanism requiring no additional training that recasts an embodied agent's observation history as a non-parametric geometric-semantic prior via a recall-ground-fuse pipeline: retrieving plausible historical frames, verifying their geometric support through alpha-compositing responsibility, and fusing only this verified evidence into the current Gaussians. On ScanNet and ScanNet++, RecallGS narrows the gap to EmbodiedSplat's full 3D branch with higher speed than the full model.

We observed that RecallGS's quality is bounded by the past observations it relies on, offering limited benefit when historical views are sparse or low-quality and additionally, grounding rasterizes only the current local Gaussians, so occluders in the global map are not modeled and visibility may be overestimated. Future work may extend retrieval with a certainty or assessment gate that evaluates the reliability of a recalled observation, for instance based on viewpoint quality, illumination, or depth confidence, before allowing it to influence the current Gaussians.

\clearpage

\bibliographystyle{splncs04}
\bibliography{reference}

\end{document}